\def\year{2020}\relax
\documentclass[letterpaper]{article} % DO NOT CHANGE THIS
\usepackage{aaai20}  % DO NOT CHANGE THIS
\usepackage{times}  % DO NOT CHANGE THIS
\usepackage{helvet} % DO NOT CHANGE THIS
\usepackage{courier}  % DO NOT CHANGE THIS
\usepackage[hyphens]{url}  % DO NOT CHANGE THIS
\usepackage{graphicx} % DO NOT CHANGE THIS
\usepackage{graphicx}  % DO NOT CHANGE THIS
\usepackage{amsmath}
\usepackage{bbold}
\usepackage{multirow}
\newcommand{\citet}[1]
{\citeauthor{#1} ̃\shortcite{#1}}
\newcommand{\citep}{\cite}

\title{MTSS: Learn from Multiple Domain Teachers and \\ Become a Multi-domain Dialogue Expert}
\author{
\Large \textbf{Shuke Peng,\textsuperscript{\rm{1,2}}~~
Feng Ji,\textsuperscript{\rm 2}~~
Zehao Lin,\textsuperscript{\rm{1,2}}~~
Shaobo Cui,\textsuperscript{\rm 2}~~
Haiqing Chen,\textsuperscript{\rm 2}~~
Yin Zhang\textsuperscript{\rm 1}\thanks{Corresponding Author: Yin Zhang, zhangyin98@zju.edu.cn}}\\ 
\textsuperscript{\rm 1}College of Computer Science and Technology, Zhejiang University\\ \textsuperscript{\rm 2}DAMO Academy, Alibaba Group\\%If you have multiple authors and multiple affiliations
\textrm{\{pengsk, georgelin, zhangyin98\}@zju.edu.cn,~~
\{zhongxiu.jf, yuanchun.csb, haiqing.chenhq\}@alibaba-inc.com} % email address must be in roman text type, not monospace or sans serif
 }
\begin{document}

\maketitle
\begin{abstract}
%\textbf{The Motivation of this paper is dialogue state representation? But after reading this paper, I am quite confused. It seems that teachers teach nothing about the dialogue state representation. They, nevertheless, impart their learnt dialogue policy by learning a better dialogue state representation in their domain. } \\
How to build a high-quality multi-domain dialogue system is a challenging work due to its  complicated and entangled dialogue state space among each domain, which seriously limits the quality of dialogue policy, and further affects the generated response. In this paper, we propose a novel method to acquire a satisfying policy and subtly circumvent the knotty dialogue state representation problem in the multi-domain setting. Inspired by real school teaching scenarios, our method is composed of multiple domain-specific teachers and a universal student.
Each individual teacher only focuses on one specific domain and learns its corresponding domain knowledge and dialogue policy based on a precisely extracted single domain dialogue state representation. Then, these domain-specific teachers impart their domain knowledge and policies to a universal student model and collectively make this student model a multi-domain dialogue expert. Experiment results show that our method reaches competitive results with SOTAs in both multi-domain and single domain setting.

\end{abstract}

\section{Introduction}
Spoken Dialogue Systems~(SDS) are widely used as assistants to help users in processing daily affairs such as booking tickets or reserving hotels. A typical dialogue system consists of three key components: spoken language understanding~(SLU), dialogue manager~(DM), and natural language generation~(NLG)\cite{maes2005conversational,maes2006system}. 
Within the procedure above, dialogue state representation is crucial since DM needs a precise representation of the present dialogue state to select an appropriate action. 
There are mainly two types of approaches for dialogue state representation: the state tracking approach and the hidden vector approach. The state tracking approach is to use a belief state tracker to extract the ontology from users' utterances~\cite{sun2014generalized,mrkvsic2016neural,zhong2018global}. Those extracted ontology, known as slots, are used as the state representation. 
The hidden vector approach, more popular utilized in end-to-end dialogue systems, is to use the hidden vector compressed from users' utterance as the state presentation~\cite{serban2016building,yao2015attention}. 
%Spoken Dialogue Systems(SDS) are widely used as assistants to help users in processing daily affairs. The architecture of a dialogue system usually consists of following key components: A Spoken Language Understanding(SLU) that understands the users' intents, a Natural Language Generator(NLG) that generates human-readable text responses, and a dialogue manager that captures the dialogue states and makes decisions for the response. 
%Tasks of the dialogue system often vary from searching for a restaurant to booking several flight tickets. The demand for finishing tasks in diverse situations requires the SDS to have the ability to handle different domains of the dialogue.
%Dialogue state representation is an essential part of building a successful dialogue system. A generally used method is to use the human-defined state representation, where the state records necessary information such as indispensable slot values the system needs. The dialogue policy then makes actions associating with the state.  In the real dialogue, a belief state tracker is usually adopted to recognize the ontology from the user's text and summarize the states by the ontology\cite{sun2014generalized,mrksivc2016neural,zhong2018global}.
%Another optional method is to use a hidden state representation. The text is compressed to hidden vectors from the raw utterance. The model summaries the context and dialogue acts are making from the hidden states\cite{serban2016building,yao2015attention}. In this setting, the dialogue system is pure an end to end model with only text as the input.
\begin{figure}[!t]
    \centering
    \includegraphics[width=0.5\linewidth]{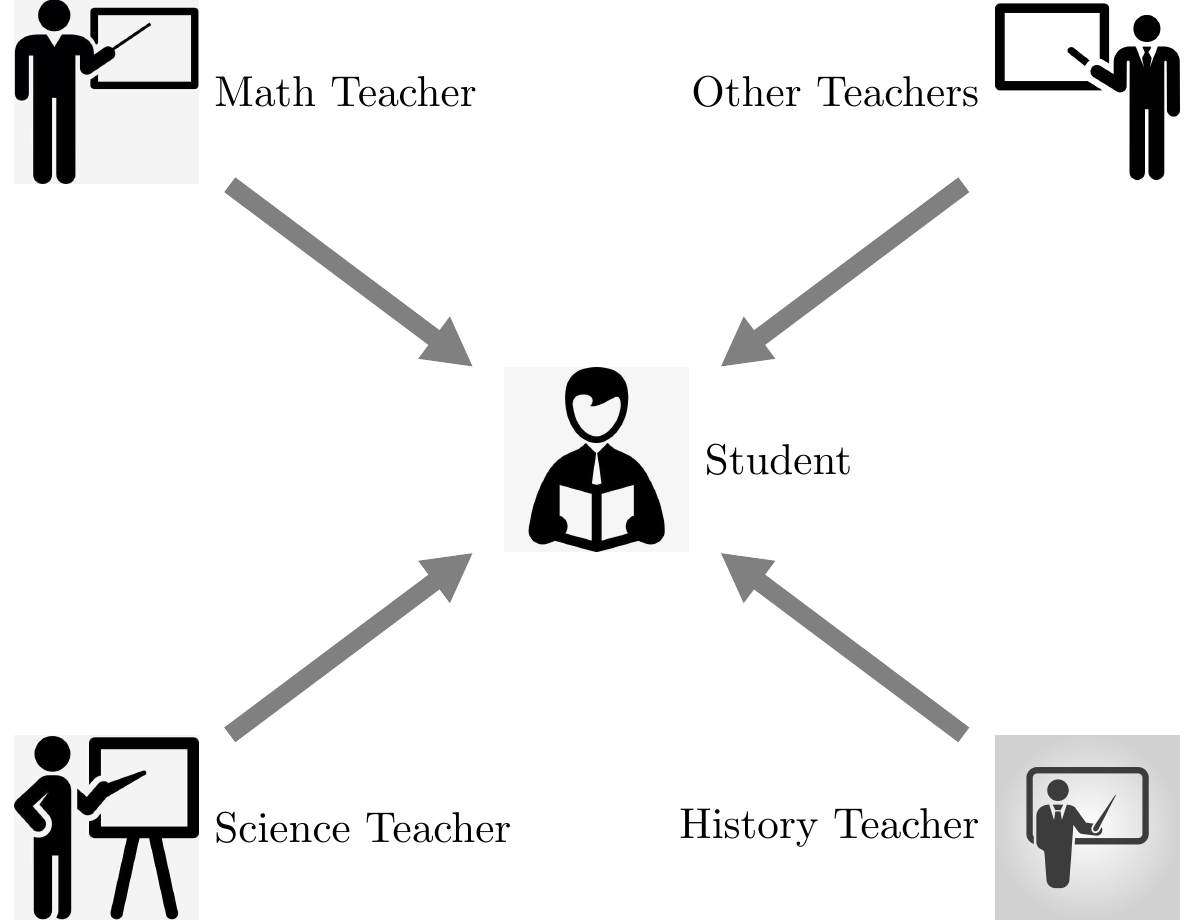}
    \caption{Learning scenarios in school}
\label{fig:teaching}
\end{figure}
The aforementioned approaches are almost satisfactory in a single domain setting dialogue task such as tickets booking since the number of the slots, and the entities are relatively small in a single domain setting. 
Nevertheless, the performance of existing dialogue state representation approaches deteriorates rapidly when it comes to multi-domain setting. For the state tracking approach, the ontology space grows enormous in multi-domain dialogue systems. This growing ontology space leads to the accuracy degeneracy of dialogue state tracking, which limits the performance of dialogue systems. As for the hidden state representation approach, the human-labelled semantic information cannot be fully used. Besides, a hidden state representation is almost a black box which makes the dialogue system incomprehensible and hard to debug. The poor-quality and inaccurate multi-domain dialogue state representation severely limits the quality of multi-domain dialogue policy and further affects the overall performance of dialogue systems. 

To build a satisfactory multi-domain dialogue system, we propose a model named Multiple Teachers Single Student~(MTSS) to subtly circumvent the complex multi-domain dialogue state representation problem and learn a quality dialogue policy in a multi-domain setting. We use multiple teacher models~(one for one domain to learn a satisfying domain-specific dialogue policy) to teach a student model to become a multi-domain dialogue expert. 
Our intuition comes from a real-life scenario in which a student has to learn many subjects such as Math, History and Science~(see Figure~\ref{fig:teaching}). 
%It is demanding for a student to tackle hard problems from different subjects by directly learning from textbooks. 
Usually, there is a full-time teacher regarding each subject. These teachers impart their professional knowledge of their respective subjects to a student. In other words, this student acquires a comprehensive understanding of all subjects by learning from these teachers. This well-educated student can achieve high performance in all subjects. 
This MTSS learning pattern is well-suited for the multi-domain dialogue systems. 
% In each domain, a specific teacher model learns more precisely about dialogue state and dialogue policy. By learning from these domain-specific teachers, a universal student model learns multi-domain knowledge and labelled semantic information. 
More specifically, \textbf{firstly}, for each domain of a multi-domain dialogue corpus, an individual teacher model is employed to learn dispersed dialogue knowledge and semantic annotations as the extra information in this single domain. Each domain teacher takes dialogue history utterances and human-labelled semantic from its corresponding domain as the dialogue state. Based on these domain-specialized dialogue state representation, these \textit{customized} teachers can acquire a high-quality dialogue policy. 
\textbf{Secondly}, these well-trained domain-specific teachers in first step \textit{impart} their learnt knowledge and dialogue policy to a universal student model through text-level guiding and policy-level guiding. 
%\textcolor{red}{I SHOULD ADD SOMETHING TO ILLUSTRATE OUR CONTRIBUTION IN POLICY LEARNING;}
We use knowledge distillation~\cite{hinton2015distilling,kim2016sequence} to implement this guiding process. 
 By learning from these domain-specific teachers, the universal student model acquires multi-domain knowledge and labelled semantic information and it finally becomes a multi-domain dialogue expert.

To sum up, the contributions are summarized as follows:
\begin{itemize}
    \item We propose a novel multi-domain dialogue system. Our model subtly circumvents the knotty multi-domain dialogue state representation problem by using multiple teacher models to learn domain-specific dialogue knowledge. With their acquired knowledge and policies, these domain-specific teacher models collectively make a single student model become a multi-domain dialogue expert. 
    % To our best knowledge, it is the first work to utilize multiple teacher models to collectively learn a comprehensive and quality dialogue multi-domain dialogue policy. 
    \item Based on MTSS, we propose a novel approach to transferring the knowledge of domain teacher models to this single student model. These teacher models guide the student model not only from the text-level but also from policy-level, which collaboratively pass the teachers' knowledge and policies to the student model. 
    %\item We use a multi-teacher single -student framework to gather the extra knowledge from individual domains to one universal model in dialogue systems.
    
    %\item Based on the extracted dialogue state representation from MTSS, we further build an end-to-end multi-domain dialogue system. 
\end{itemize}

\section{Related work} \label{sec:literature}
%\textbf{I think this section is too discursive. I suppose it maybe better if we separate this section into several subsubsection and focus only one topic in each subsubsection -- \textit{FROM SHAOBO}}
\subsubsection{Multi-domain dialogue systems}
% The dialogue systems achieve excellent results in limited domain dialogues. When it comes to the multi-domain situation, the problem becomes more complex and challengeable. \citet{budzianowski2018multiwoz} recently released a multi-domain human-human conversation corpus. The dataset contains up to 7 domains, a large number of entity slots and dialogue acts. The topic and the user goal in every conversation often involve more than one domain. 
Recently, multi-domain dialogue systems have attracted increasing attention.  
%because their wide applications 
%\textcolor{red}{such as dynamic dialogue processing for users as a daily assistant\cite{pakucs2003towards}, the scalable action spaces for a dialogue system to share knowledge between domains\cite{dzikovska2003integrating}. 
The rule-based multi-domain dialogue systems~\cite{pakucs2003towards} are faced with the insufficiency of the scalability. 
With the development of deep learning, some multi-domain dialogue systems models are proposed based on neural network~\cite{wen2016multi,ultes2017pydial}.
%There are some works focusing on multi-domain dialogue systems. 
%As the neural network is widely used in dialogue systems, ideas appeared to handle the multi-domain dialogue with a deep network\cite{wen2016multi,ultes2017pydial}.  
% Recently, some state-of-the-art works have been done to build a multi-domain dialogue system. 
\citet{zhao2019rethinking}  propose the Latent Action Reinforcement Learning~(LaRL) model, which uses reinforcement learning to train a policy module to select the best latent action.
The Hierarchical Disentangled Self-Attention~(HDSA)~\cite{chen2019semantically} model uses hierarchical dialogue act representation to deal with the large size of dialogue acts. 
%It utilizes a disentangled self-attention architecture to control the information flow by the hierarchical dialogue acts. 
Both two works were applied in the MultiWOZ~\cite{budzianowski2018multiwoz} dataset and achieved excellent results.
%However, both models take the human-labelled states as part of the input for granted. Practically, the external state tracker has significant effects on the models' performance. The models mentioned above have ignored this. To overcome such a shortage, our model applies a text-in, text-out framework, and gets better results in the MultiWOZ dataset.
\subsubsection{The representation of dialogue states}
  % The dialogue states can be regarded as a Partially Observable Markov Decision Processes(POMDPs)\cite{thomson2010bayesian}. 
  %The features often contain slots that must be filled in the task, and domain tags if there is more than one domain. States generated by this kind of embedding method are often well explainable.
 A commonly-used approach to representing dialogue states is to use the multi-hot embedding vector of human-defined features as the state representation. 
 This type of approach needs an external dialogue state tracker to recognize correct features from users' utterance. Many works have been done on this issue, such as a rule-based state tracker~\cite{sun2014generalized} or a Neural Belief Tracker~(NBT)~\cite{mrkvsic2016neural}. Some works are focusing on state trackers that track user intent and slot values in multi-domain settings~\cite{rastogi2017scalable,goel2018flexible}. %(To be extended)
 In addition to using human-defined features as the dialogue state representation, another approach is to use the hidden state vector generated directly from the raw text as the state representation. 
 Without handcrafted features, Hierarchical Recurrent Encoder-Decoder~(HRED) based dialogue systems\cite{sordoni2015hierarchical,serban2016building,serban2017hierarchical} encode the dialogue history into a hidden vector to represent the current dialogue state in open-domain dialogue systems. 
\begin{figure*}[ht]
    \centering
    \includegraphics[width=0.8\linewidth]{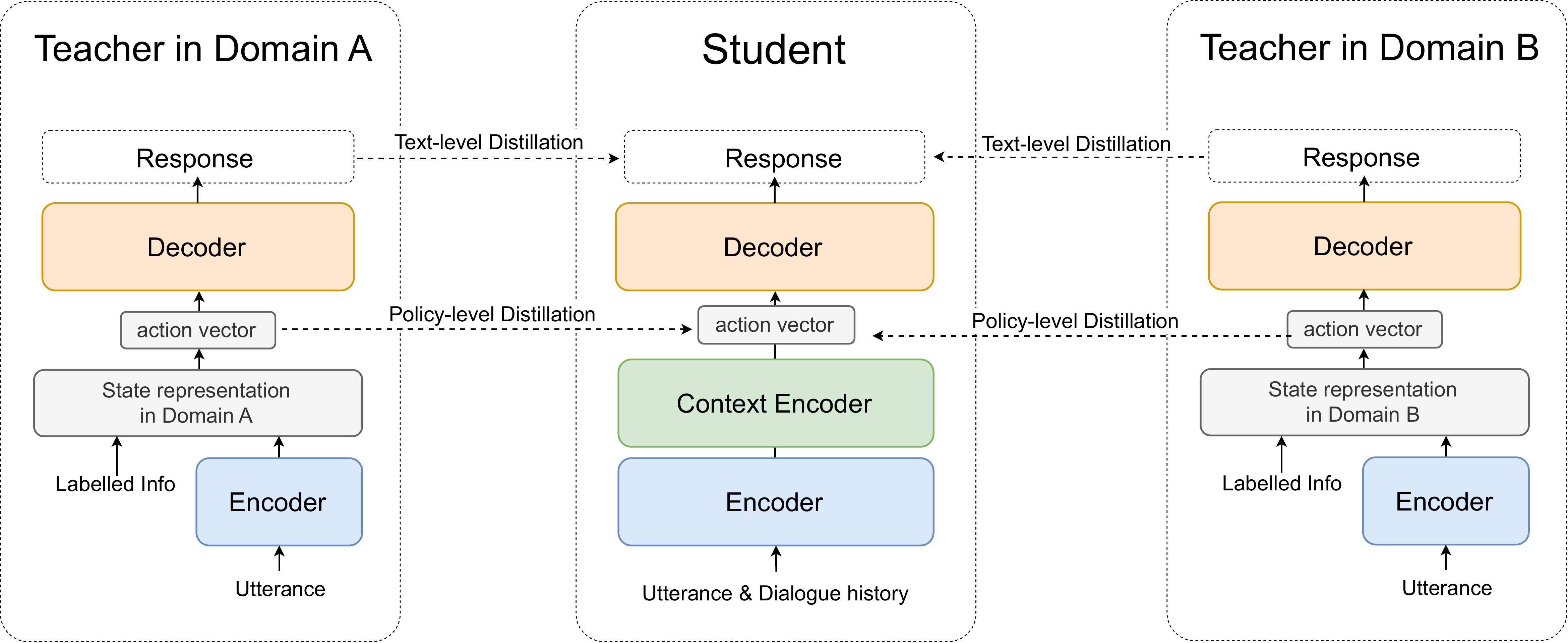}
    \caption{The teacher-student framework that transfers the knowledge from teachers to the student.}
    \label{fig:TS}
\end{figure*}
\subsubsection{The Teacher-student Framework}
 The teacher-student framework was first applied in the neural network by \citet{hinton2015distilling} in the knowledge distillation approach. 
 In the teacher-student framework, a massive teacher model transfers their knowledge to a much smaller student model or several assembled teacher models collectively transfer their knowledge to a student model.
 %, or several assembled teacher models integrate to one student. 
 %Earliest applications of knowledge distillation are mainly in computer vision. 
 Recent works show that knowledge distillation based teacher-student method works well in a language model~\cite{kim2016sequence}. \citet{tan2019multilingual} proposed a multi-teacher single-student architecture to solve the multilingual neural machine translation problem. Individual models are built as teachers, and the multilingual model is trained to fit both the ground truth and the outputs of individual models simultaneously through knowledge distillation. In this way, the student model can reach comparable or even better accuracy in each language pair than these teacher models. Our work adopts a similar architecture, but we focus on multi-domain dialogue systems, which is more challenging since it involves complicated multi-domain dialogue policy learning. 
% adapts the multi-teacher single-student in the multi-domain dialogues. 
%We also apply the policy distillation\cite{rusu2015policy} to our model. The original usage of a policy distillation is to transfer the policy from a well-trained agent to a new model in reinforcement learning. We use the method in the teacher-student framework to gain more knowledge from teachers as possible.

\section{The Framework of Multiple Teachers Single Student~(MTSS) Model} \label{sec:model}
In this section, we present the framework of our proposed Multiple Teachers Single Student~(MTSS) model in Section~\ref{sec:model:overview} and detail the teacher and the student component in Section~\ref{sec:model:teacher} and Section~\ref{sec:model:student} respectively. We leave how the multiple teacher models impart their acquired knowledge to the student model in Section~\ref{sec:learn}. 

\subsection{The Overview of MTSS}\label{sec:model:overview}
The overview of MTSS is presented in Figure~\ref{fig:TS}~(For a clear illustration, we only plot two teacher models in the figure, which is sufficient to illustrate the whole framework and the working procedure). MTSS consists of two types of components: the student model and the teacher model. 
There are $N$ teacher models and one single student in MTSS, where $N$ is the number of dialogue corpus domain. In other words, each teacher model in MTSS is associated with one domain of the dialogue corpus. 
In the training phase, the teacher model and the student model are trained with different input: 
\begin{itemize}
\item The teacher takes the utterance and the human-labelled states as the input.  The states labelled by human are of the highest accuracy, provide the teacher model sufficient information in dialogue policy decision and responding. 
\item The student takes the utterance and the history dialogues as the input. 
% The raw context is harder to comprehend for the model comparing with the well-formed state representation. 
\end{itemize}

These well-trained teacher model impart their knowledge in both text-level and policy-level. The text-level guidance is to make the student generate a similar response as the teacher models while the policy-level is to make the student learn the policies of these teachers, which make sure the student model can fully \textit{assimilate} the knowledge of teachers. We will introduce the details of interactions between the student model and teacher models in Section~\ref{sec:learn}. 

After the training phase, the student model has acquired sufficient multi-domain knowledge and a satisfying multi-domain dialogue policy.
At the testing phase, the student model only takes raw context utterances as input and can generate high-quality responses. %It can get rid of a state tracker and performs well with the raw context input. 
% Our multi-domain dialogue generation system can be illustrated as three parts: A multi-domain hierarchical dialogue generation model, serving as the primary model to learn external knowledge. Several individual dialogue models play the roles of teacher models to guide a student. And a guiding step that transfers the knowledge from individual models to the universal one. In our model, the state $s_t$ for the teacher model is directly from human labeled semantic in the utterance, and the student model generates the $s_t$ itself from all passed user utterance. The detail of the two models will be described in the rest of this section.

\subsection{Multiple Teachers: One Teacher for One Domain}
\label{sec:model:teacher}
% We train dialogue models as teachers in each domain. Differ from the universal model, and the individual models take human-labelled semantic as the states of the conversation. The states represent the necessary information in the dialogue history. It's no need for a teacher to model the whole conversation text. So we use a simple seq2seq model alone with state inputs as the teacher model.
\subsubsection{The structure of the teacher model}
\begin{figure}[!t] 
    \centering
    \includegraphics[width=0.4\linewidth]{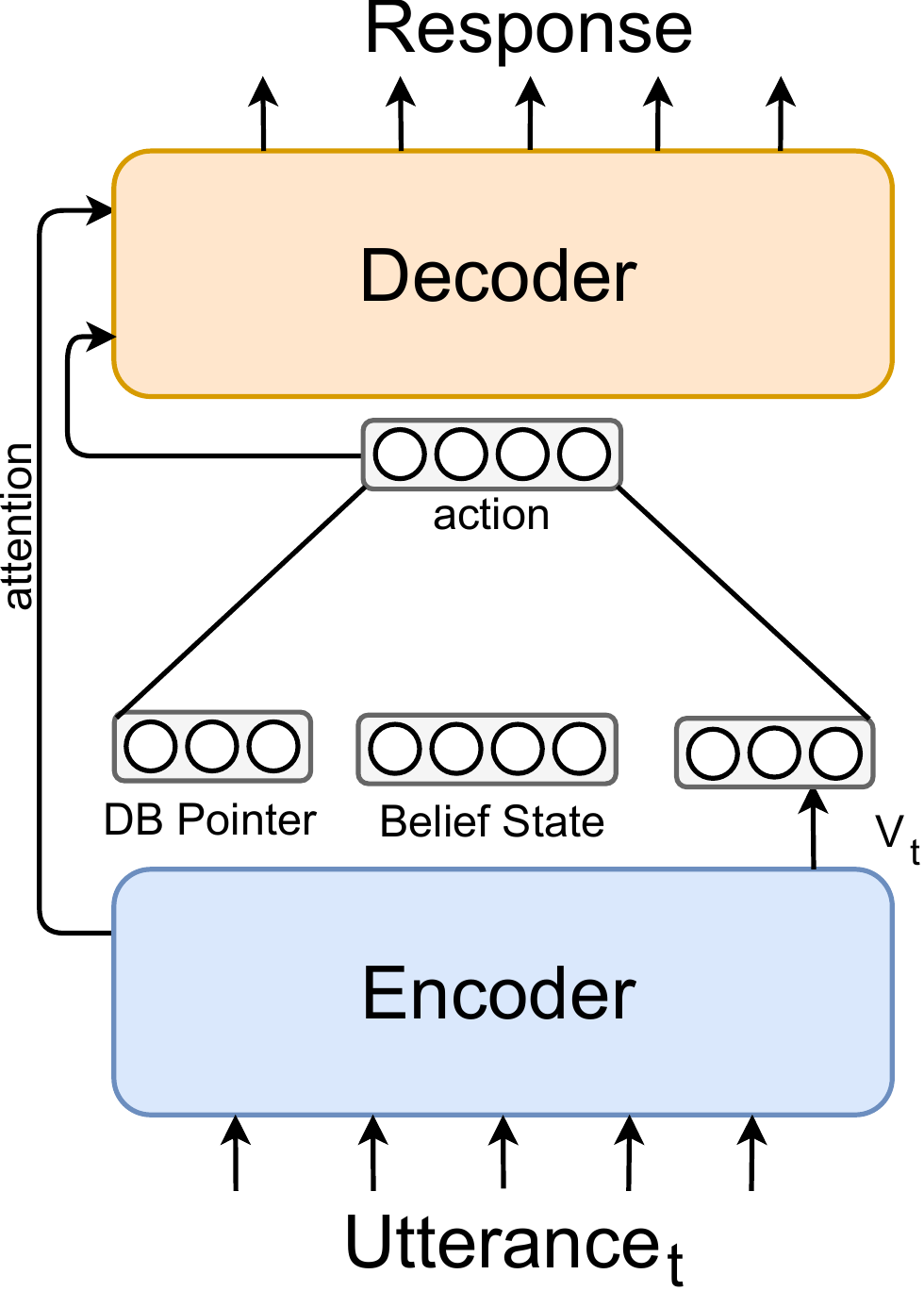}
    \caption{The teacher model pre-trained from each domain}
    \label{fig:baselines}
\end{figure}

We adopt \citet{budzianowski2018multiwoz} as the basic structure. As shown in Figure~\ref{fig:baselines}, it contains three parts: the encoder, the decoder and a middle policy model that takes both the utterance representation $u_t$ as well as the human-defined feature $e_t$ as the input. 
The feature consists of two vector representations. The first part is the belief state vector $\mathbf{v}_\mathrm b$, where each dimension of the vector stands for the one-hot value of a specific slot in each  domain, a slot value receiving from the user. If the slot value appears, the corresponding value in the vector is set to 1. Otherwise, the value is 0. Thus all values of $\mathbf{v}_\mathrm b$ stand for necessary information the system keep at the current state. At every turn, the belief state is updated according to the semantic labelling of the users' utterances. Another construct of the state is the database pointer vector $\mathbf{v}_\mathrm{kb}$, where a database pointer vector stands for the number of the corresponding entities that match the request of the user. We use a 4-dimensional one-hot embedding vector, and each position embedding means separately 0, 1, 2 and more than 3 candidate entities. We concatenate three vectors: the utterance vector $\mathbf{v}^u_t$, the belief state $\mathbf{v}_\mathrm{b}$, and the database pointer $\mathbf{v}_\mathrm{kb}$, to get the vector of the current state $s_t$ in the conversation.

Then we feed the concatenated vector to the policy model. The vector is processed with a nonlinear layer with $\tanh$ as the activation function, and the action vector $\mathbf{a}_t$ is generated from this layer:
\begin{equation*}
\mathbf{a}_t = \mathrm{tanh}(\mathbf{w}\cdot[\mathbf{v}^u_t;\mathbf{v}_\mathrm{b};\mathbf{v}_\mathrm{kb}]),
\end{equation*}
where $[;]$ stands for concatenation. The action $\mathbf{a}_t$ is finally delivered to the decoder module and the response is generated with an addition of the attention mechanism. 
We train teacher models individually in each domain. Thus the meaning of the belief state differs in teachers. After the teachers are well pre-trained in all domains, we take the teachers as the guidance to train the student model using the teacher-student framework.

\subsubsection{Training of the teacher model}
The teacher model directly learns from the ground truth.
For a teacher model, given the user utterance $u$ and the state representation $s$, the purpose of the model is to minimize the negative log likelihood loss between the generated response $\hat r$ with a ground truth response $r=\{w^r_0, w^r_1, ..., w^r_m\}$. That can be written as:
\begin{equation}
\begin{split}
J_\text{NLL}(\hat r|u, s)=&\\
-\sum_{i=0}^m\sum_{\hat w_i\in \mathcal V}\mathbb 1&\{\hat w_i=w^r_i\}\log p(\hat w_i|u,s,w^r_{0\sim i-1};\phi),
\end{split}
\end{equation}
where the $\mathcal{V}$ is the vocabulary of all possible words, $\phi$ is the parameters of the teacher model and the symbol $\mathbb 1\{\cdot\}$ stands for the indicator function.

\subsection{Single Student: A Universal Multi-domain Dialogue System} \label{sec:model:student}
\subsubsection{The structure of the student model}
\begin{figure}[!t]
    \centering
    \includegraphics[width=0.4\linewidth]{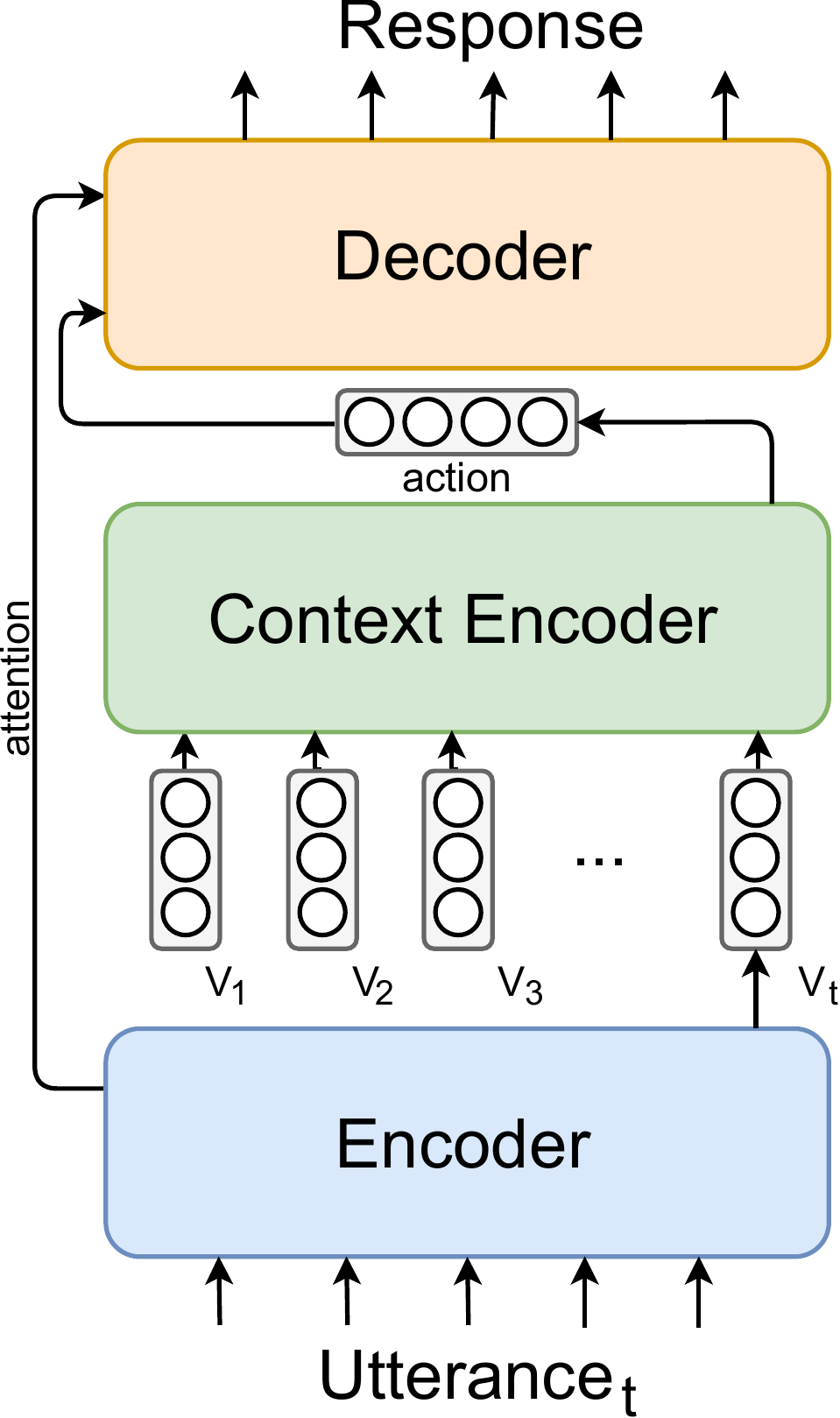}
    \caption{The student model architecture}
    \label{fig:HRED}
\end{figure}

The universal dialogue system, also the student model is the final produced model of our framework. The universal model takes no extra state information as the input. And it should have the ability to model the whole context, summarize the history states directly from the text. Under such consideration, we adopt the HRED~\cite{sordoni2015hierarchical,serban2016building} model as our universal dialogue system's base architecture.
We use an encoder module to encode the user utterance to a latent vector representation and summarize all utterances' vectors with a context-level  encoder in hierarchical encoder-decoder architecture, as shown in Figure \ref{fig:HRED}. 
At the time $t$, for an utterance $u_t$ contains $m$ words $(\mathbf{w}_0, \mathbf{w}_1, ..., \mathbf{w}_m)$. The encoder is an LSTM~\cite{hochreiter1997long} network:
\begin{equation*}
\mathbf{h}_t=\mathbf{v}_{tm}^w=\mathrm{LSTM}_e(\mathbf{h}_0;\mathbf{w}_{t0},\mathbf{w}_{t1},...\mathbf{w}_{tm}),
\end{equation*}
Then we consider the last hidden state of the LSTM as the utterance representation vector $\mathbf{v}^u_t=\mathbf{h}_t$, and take the hierarchical encoder as the context-level policy module. The action $\mathbf{a}_t$ is made based on the all history utterances. We use another LSTM as the context-level encoder:
$$\mathbf{a}_t=\mathrm{LSTM}_c(\mathbf{v}^u_0, \mathbf{v}^u_1, ..., \mathbf{v}^u_t)$$ The action $\mathbf{a}_t$ is in the form of an abstract latent vector, serving as the guidance for the dialogue system to make proper responses. By regarding the context-encoder output as the action representation, we'll see how this representation facilitates the performance of our model using the teacher-student framework. 

The action is fed into the generation part lately. The NLG module regards the action as the initial state of LSTM and generates the final response $r_t$. With the addition of the attention mechanism, the decoder model can be written as:
\begin{equation*}
    \mathbf{v}^r_i=\mathrm{LSTM}_d(\mathbf{a}_t,\mathbf{v}^w_{0\sim m},\mathbf{v}^r_{0\sim i-1}),
\end{equation*}
where $\mathbf{v}^w_j$ is the output of the encoder in the position of the $j$-th word . 
\subsubsection{The guidance from ground truth for the student model}
Same as the training process of teacher model, the student model learns for the ground truth too. 
In contrast to the input for a teacher model, there is no 
explicit state representation as an input for the student. Instead, the student needs to summarize the hidden state from the context input itself.
In addition to the guidance from the ground truth, the student model also learns from domain teachers, which will be elaborated in Section~\ref{sec:learn}. 

\section{How Does The Single Student Learn from Domain Teachers?} \label{sec:learn}
%In this section, we elaborate the guidance for the optimization of the student model. There are mainly two types of guidance: the guidance from the ground truth~(Section~\ref{sec:learn:gt}) and the guidance from domain teachers~(Section~\ref{sec:learn:teacher}). 
In this section, we elaborate on the methods of transferring the knowledge from domain teachers to the student model. This transferring process can also be viewed as knowledge distillation~\cite{hinton2015distilling,kim2016sequence} from teacher models to the single student model. These domain-specific teachers cooperatively guide the student model from both text-level~(Section~\ref{sec:learn:text}) and policy-level~(Section~\ref{sec:learn:policy}), which makes sure the student can fully absorb the knowledge of these domain-specific teachers.

\subsection{Text-level Guiding} \label{sec:learn:text}
We expect that the student should output a similar response as the teachers do. At each timestep, the student model is expected to generate the same output distribution as the teachers do. To enforce this objective, we use the cross entropy loss to measure the probability similarity between the output distributions of student and the teachers. The loss of the text-level distillation is:
\begin{equation} \label{eq:jkd}
\begin{split}
 J_\text{KD}=-\sum^m_{i=0}\sum_{w^r_i\in \mathcal V}p(w^r_i|u,s, w^r_{0\sim i-1};\phi)\\\log{p(w^r_i|u,c, w^r_{0\sim i-1};\theta)},
\end{split}
\end{equation}
in which $\phi$ is the parameter of the teacher models and $\theta$ is the parameter of the student model.
%And $\mathcal W$ is the word range the student learns a distribution from teachers. Notice that we have different choices for $\mathcal{W}$ in Equation~\ref{eq:jkd}: 
And $\mathcal V$ is the whole vocabulary. 
For the grounding truth of the training data, the generation part of the model learns only the one-hot value at each position. For text-level distillation, the guidance from the teachers' output applies a smoother distribution of the probability of words. The distillation brings naturalness and correctness for the dialogue generation.
%\begin{enumerate}
    %\item \textbf{Vocabulary size distillation}: all the output logits at each position are used for knowledge distillation. The word range $\mathcal W$ is equal to the vocabulary $\mathcal V$. The vocabulary size distillation is the naive way to distil the knowledge from the teacher model. For the grounding truth of the training data, the generation part of the model learns only the one-hot value at each position. For the distillation, the guidance from the teachers' output applies a smoother distribution of the probability of words. The vocabulary size distillation brings fluency and correctness for the dialogue generation.
    %\item \textbf{Top-K distillation}: only top k logits at each position are used for knowledge distillation\cite{tan2019multilingual}. The word range $\mathcal W$ contains k words that have the highest output probability. This method of distillation doesn't use the full probability of the vocabulary size at every step of response generation. The top k of the logits are selected in the teachers' output to be the guidance of the student's training. This kind method of distillation is more efficient than a full vocabulary one. Not all of the words should be taken into consideration in the generation of the sentence, so the omitting of the low probability words helps in the guiding process.
%\end{enumerate}
% \texthctbf{Sequence-level distillation}. Also called beam distillation\cite{kim2016sequence}.
\begin{center}
\small\addtolength{\tabcolsep}{-2pt}
\begin{table*}[!ht]
\centering
\begin{tabular}{|c|cc|cc|cc|cc|cc|cc|}
    \hline
    \multirow{2}{*}{\textbf{Models}}& \multicolumn{2}{c|}{\textbf{Restaurant}}
    & \multicolumn{2}{c|}{\textbf{Hotel}}
    & \multicolumn{2}{c|}{\textbf{Train}}
    & \multicolumn{2}{c|}{\textbf{Attraction}}
    & \multicolumn{2}{c|}{\textbf{Taxi}}
    & \multicolumn{2}{c|}{\textbf{General}} \\
    & \textbf{BLEU} &\textbf{ER} & \textbf{BLEU} &\textbf{ER}& \textbf{BLEU} &\textbf{ER}& \textbf{BLEU} &\textbf{ER}& \textbf{BLEU} &\textbf{ER}& \textbf{BLEU} &\textbf{ER}\\ 
    \hline
    \hline
    \multicolumn{13}{|c|}{\textit{Teachers}} \\ \hline
     Universal teacher & 16.5 &\textbf{69.89} & 14.1 &52.52 & 22.3 &\textbf{63.19} & 13.1 &58.96 & 15.7 &48.03 & 19.8 &- \\
     Individual teachers & \textbf{20.5} &68.60 & \textbf{16.4} &\textbf{56.43} & \textbf{23.1} &60.31 & \textbf{16.6} &\textbf{67.65} & \textbf{17.7} &\textbf{86.68} & \textbf{23.0} &-  \\
     \hline
     \hline
    \multicolumn{13}{|c|}{\textit{Students}}\\ \hline
    HRED(No teacher) & 17.1 & \textbf{54.82} & 15.0 & 44.95 & 17.2 & 47.27 & \textbf{16.8} & \textbf{71.78} & 15.5 & \textbf{76.64} & \textbf{22.7} & - \\
     HRED-MTSS & \textbf{18.1} & 50.89 & \textbf{16.5} & \textbf{45.91} & \textbf{19.9} & \textbf{56.19} & 16.3 & 66.82 & \textbf{16.4} & 64.85 & 19.9 & -   \\ \hline
\end{tabular}
\caption{Performance of different teachers
and different students 
in each domain.
A universal multi-domain teacher model trains on the whole dataset and several individual teacher models train in each domain. ER: entity recall.}
\label{tbl:teacher}
\end{table*}
\end{center}
\begin{table}[!ht]
    \centering
    \begin{tabular}{|c|cc|}
        \hline
         \multirow{2}{*}{\textbf{Domain}} & \multicolumn{2}{c|}{\textbf{Number of Turns}}  \\
         \cline{2-3}
         & \textbf{Train} & \textbf{Test} \\
         \hline
         \hline
        Restaurant &  13471 & 1571 \\
        Hotel & 12943 & 1506 \\
        Train & 10612 & 1735\\
        Attraction & 7054 & 1061 \\
        Taxi & 2996 & 419 \\
        Hospital & 593 & 0 \\
        Police & 463 & 4 \\
        General & 8646 & 1072 \\ \hline
    \end{tabular}
    \caption{Number of turns in each domain when MultiWOZ is split.}
    \label{tbl:domain_num}
\end{table}
\subsection{Policy-level Guiding} \label{sec:learn:policy}
We also expect that the universal model can acquire the dialogue policies of these teachers. In other words, we expect that the teacher models and the student model should have similar action vector if provided with similar input.
%Though the teacher model and the student model differ in the structure of the policy part, they have the same decoder module as the NLG part. 
%Thus the teaching process of the policy part from teacher models is helpful to a student model. 
We use the action $\mathbf a^\text T$ from the teachers' policy output as the extra information to train the student's policy. For $\mathbf a^\text T$ and $\mathbf a^\text S$ are both in the form of latent vectors. In the training phase, we use mean squared error~(MSE) loss to force the student to learn the policies of the teachers:
\begin{equation}
    J_{\text {KD}-\pi}=\sum^k_{i=0}(a^\text T_i-a^\text S_i)^2,
\end{equation}

We use both the ground truth~(Section~\ref{sec:model:student}) and the teachers' guidance as the training target. We add the text-level distillation loss and the policy-level distillation loss to the loss of the ground truth. To adjust the effect of the teachers and balance the weights of the different losses, we apply a weight scalar $\alpha_1$ to the text-level distillation loss and another weight scalar $\alpha_2$ to the policy-level one. Finally, the combination training loss $J_\theta$ of the student model can be illustrated as: 
%\textbf{We use both the grounding truth~(Section~\ref{sec:model:student}) and the teachers' output as the target data in the training phase, and apply the negative of the log-likelihood as the loss of output distillation. We add this loss to the loss of the grounding truth. To adjust the effect of the teachers, we apply a weighted scalar $\alpha_1$ to change the importance of teachers while training.}

%\textbf{We add the policy distillation loss $J_{\text {KD}-\pi}$ to the existing loss by multiplying another weighted scalar $\alpha_2$ as the text-level distillation does.}
\begin{equation} \label{eq:all}
    J_{\theta} = J_\text{NLL}+\alpha_1 J_\text{KD}+\alpha_2 J_{\text{KD}-\pi},
\end{equation}
then we train the student model to minimize the combination loss $J_\theta$ to implement the guiding of teacher models.

%\textbf{Add an equation to further illustrate combination of loss}
\section{Experiments} \label{sec:experiments}
%\textbf{Please Reorganize the structure of this section}

%To figure out the performance of the approaches mentioned above, we apply our model to a multi-domain dialogue problem to test the ability of our teacher-student based dialogue systems. We compare our model with state-of-the-art methods.
In this section, we elaborate the experiment settings~(Section~\ref{sec:experiments:settings}), the baselines we use~(Section~\ref{sec:experiments:baselines}), and the analysis of experimental results~(Section~\ref{sec:experiments:results}). 
\begin{table}[!ht]
\centering
\begin{tabular}{|c|ccc|}
     \hline
     \multirow{2}{*}{\textbf{Models}}& \multicolumn{3}{c|}{\textbf{Multi-domain}}
     \\
     & \textbf{BLEU} & \textbf{Inform} & \textbf{Success} \\
     \hline
     \hline
     \multicolumn{4}{|c|}{\textit{Comparisons}}\\ \hline
     Seq2seq & 16.7 & 65.7 & 44.4   \\
     HRED & 17.5 & 70.7  & 60.9    \\
     Seq2seq + MDBT & 13.1 & 69.3  & 30.0   \\
     Seq2seq + TRADE & 13.2 & 65.9 & 34.6 \\
     HRED + MDBT & 13.1 & 68.8 & 35.5
     \\
     HRED + TRADE & 13.7 & 70.8 & 41.8 \\
     HRED-MTSS(ours) & \textbf{ 18.7} & \textbf{77.5} & \textbf{64.9} \\
     \hline
     \hline
     \multicolumn{4}{|c|}{\textit{State-of-the-art models}}\\\hline
     LaRL + TRADE & 12.4 & \textbf{79.5} & 44.7 \\
     HDSA + TRADE & \textbf{20.1} & 76.4 & \textbf{65.9} \\
     \hline
     \hline
     \multicolumn{4}{|c|}{\textit{Models with manual states}}\\ \hline
     Seq2seq + Manual states & 17.8 & 75.4 & 62.8 \\
     HRED + Manual states & 19.3  & 75.2 & 66.2     \\
     HDSA + Manual states & \textbf{22.9} & \textbf{82.3} & \textbf{75.1} \\
     \hline
\end{tabular}
\caption{Performance on the multi-domain environment.}
%between the raw models, the models use state trackers and the model uses knowledge distillation. The models with Manual states take the human labeled states as input and set as the upper bound of our metrics.}
\label{tbl:multi}
\end{table}

\begin{center}
\small\addtolength{\tabcolsep}{-2pt}
\begin{table*}[!ht]
    \centering
    \begin{tabular}{|c|cc|cc|cc|cc|}
    \hline
    \multirow{2}{*}{\textbf{Models}}& \multicolumn{2}{c|}{\textbf{Restaurant}}
    & \multicolumn{2}{c|}{\textbf{Hotel}}
    & \multicolumn{2}{c|}{\textbf{Train}} 
    & \multicolumn{2}{c|}{\textbf{Attraction}}\\
    & \textbf{Inform} & \textbf{Success} & \textbf{Inform} & \textbf{Success} & \textbf{Inform} & \textbf{Success} & \textbf{Inform} & \textbf{Success} \\
    \hline
    \hline
        Seq2seq + TRADE & 88.6 & 57.9 & 90.9 & 42.4  & 72.1 & 60.8 & 63.9 & 55.3   \\
        HRED + TRADE & \textbf{91.8} & 74.4 & 81.7 & 50.5 & 76.2 & 62.6 & 76.8 & 65.4   \\
        HDSA + TRADE & 78.5 & 68.6 & \textbf{91.4} & \textbf{85.3} & 81.4 & 80.4 & \textbf{93.9} & \textbf{82.1}   \\
         HRED-MTSS(ours) & 87.4 & \textbf{81.2} & 86.8 & 81.5 & \textbf{85.1} & \textbf{83.4} & 86.6 & 74.5   \\
         \hline
    \end{tabular}
    \caption{Results on different domains}
    %of our model comparing with the SOTA models}
    \label{tbl:domains}
\end{table*}
\end{center}
\subsection{Experiment Settings} \label{sec:experiments:settings}
\subsubsection{Dataset}
We choose MultiWOZ~\cite{budzianowski2018multiwoz}, a multi-domain human-human conversation corpus, as our dataset. The MultiWOZ dataset consists of dialogue turns in 7 domains, respectively including restaurant, hotel, attraction, taxi, train, hospital and police. The conversation in MultiWOZ aims at satisfying users‘ intents, and informs the necessary information the user needs about some entities. An episode of conversation contains around 14 turns of dialogues between the user and the system. Several episodes' topics are limited in one domain from beginning to the end turn, while others' are switching among the conversation in 2 to up 5 domains. In each domain, there are about 4 slots that the system can receive from the user and about 3 properties of the entity the system should provide to the user. For example, in a restaurant domain, the user can choose the area, the price range and the food type of a restaurant, and the information the system should offer about the restaurant includes the address, the reference number, the phone number and other essential properties.

To test the response quality of the models, we take a pre-processing on the dataset: we replace the names of the entities and their property values with placeholders. Then we manually generate the belief states and the database pointers, as the extra inputs of teachers, from the human labelled semantics.
%\textcolor{blue}{HARD TO FOLLOW! We split the dataset into 7 specific domains by tagging the domain of each turn by the entities mentioned by users. There are some turns that cannot be tagged as any of 7 domains, so we allocate these turns to a generic domain. Then  }
All the dialogue turns are split to 7 specific domains based on the domain tags, which are given by MultiWOZ dataset and are determined by entities in the dialogue turns. For the dialogue turns that don't belong to these 7 domains, they are included into a generic domain. In other words, we have 8 separate dialogue turn sets, each set corresponds to an individual domain. We train 8 individual teachers for each domain.
Table~\ref{tbl:domain_num} shows the number of training and testing turns in each domain after the dataset is split.
%\textcolor{red}{Besides, we use the delexicalization as a pre-processing method for the dialogue text. All the slot values are replaced with placeholders. The delexicalization is a standard approach for the MultiWOZ dataset.}
Besides, following the pre-processing instruction of MultiWOZ, all dialogue turns are delexicalized, which means all the slot values are replaced with placeholders.

% Though the fluency of the conversation may be affected by the lacking of the context in training individual teachers, we think the teacher models can be well trained as they take the turn-level sentences as input as well as the manual state.

\subsubsection{Experiment Settings}
We construct two vocabularies from the dataset, the input one and the output one. For the input vocabulary, we discard the words appear less than 5 times. About 1300 words remain in input vocabulary. For the output vocabulary, we limited the size to 500. We use two types of embeddings for the input and  the output vocabularies. The embedding size is set to 50. 
The hidden layer size of LSTM layers in all involved models is set to 150. The teacher models are the Seq2seq architecture, the encoder and the decoder are 150 dimensions hidden layer of LSTM networks as well. For each teacher model, we trained it on its respective domain, and find the model which has the best entity matching recall rate as the guidance. For the student model, we use Adam optimizer, and the learning rate is 0.005. As for $\alpha_1$ and $\alpha_2$ in Equation~\ref{eq:all}, both $\alpha_1$ and $\alpha_2$ are set to 0.005 for balancing the guidance from the ground truth and the teacher models.  
To test the stability and get reliable results, we repeat each experiment setting 3 times and some of them for 5 times. 

\subsubsection{Training Strategies}
% After the dataset is split, we construct teacher models at each domain. The simplest way to build a domain teacher is to train the model with the domain data from random parameters. We consider this way a cold-start. However, the cold-start method fails at several domains for lack of data. E.g., 
In the training phase of the teacher models, we found that the sub-dataset of some domains are limited. For instance, the sub-dataset of the police domain only accounts for 0.82\% of all training data, which results in poor performance of these teacher models. To solve this problem, we use a warm-start strategy: we use a pre-trained model $T_\text{all}$ trained on the whole the training dataset as the starts, and each teacher model is fine-tuned from $T_\text{all}$. This warm-up strategy ensures the domain-specific teachers have equal or higher performance than $T_\text{all}$.

% In the training teacher models, we found that some domains are lack of data. E.g., the data of the police domain only accounts for 0.82\% of all training data. Insufficient data leads to a poor teacher in the domain. To overcome such a shortage, we use a pre-trained universal model as the starts, and each teacher model is fine-tuned from the universal model. This step ensures the teachers have equal or higher performance than the universal one. 
% The data has been split to turn-level, so the frequently used episode-level metrics for dialogue are not suitable. To measure the teachers' abilities, we take two kinds of turn-level metrics into consideration: the BLEU score and the Entity Matching Recall(EMR), the EMR rate describes the ratio that whether the generated responses cover entities mentioned by the ground truth responses. We takes both two measures into consideration during training the best teachers.

\subsubsection{Evaluation Metrics}
To measure the performance of different models, we use several examined metrics to evaluate the generated response. 
\begin{enumerate}
\item {BLEU}: we calculate BLEU-4~\cite{papineni2002bleu} scores to measure the similarity between the real response and the generated one. 
% For BLEU scores show less correlation about the quality of the dialogue content, we apply other measurements. 
\item {Inform rate and Success rate}: We use two metrics that are suggested by \citet{budzianowski2018multiwoz}, as the estimations for the MultiWOZ dataset in the dialogue context to text task. Both the measurements are on the episode-level. The Inform rate indicates whether the dialogue system suggests suitable entities according to the user's intent in an episode. The Success rate illustrates if the system provides all the correct properties for the user requests after a success informing. 
\item {Entity Recall}: Entity Recall~(ER) measures the recall score of the entities between the generated response and the ground truth. ER is a turn-level metrics and used to evaluate the performances of the teachers.

%\textcolor{red}{THE FOLLOWING CONTENT SHOULD NOT BE EVALUATION METRICS} We run the models on the test dataset, which includes 1000 episodes of conversations, then count out the ratios of the successfully informed dialogues and successful ones. We also test the performance of our multi-domain model in the respective domain. We choose the vastest 4 domains within all 7 - the restaurant, the hotel, the attraction, and the train, testing in the domains to show more detail measurements of our model comparing with others.
\end{enumerate}

\subsection{Baselines} \label{sec:experiments:baselines}
%\textbf{Please Rewrite this subsection and itemize the baselines methods. Additionally, Baselines should be a better name for this subsection. Please take into consideration and discuss with me. -- \textit{FROM SHAOBO}}

\begin{itemize}
    %\item\textbf{Seq2seq \& HRED} Those are the two base end to end dialogue models, implemented with a Seq2seq architecture and a HRED architecture individually. 
    \item \textbf{Seq2Seq}: the vanilla Seq2Seq model~\cite{cho2014learning}.
    \item \textbf{HRED}: the HRED architecture proposed in \citet{sordoni2015hierarchical}.
    \item \textbf{Seq2Seq + MDBT}: the Seq2Seq model with the Multi-domain Belief Tracker~(MDBT)~\cite{ramadan2018large} as the state tracking model.
    \item  \textbf{Seq2Seq + TRADE}: the Seq2Seq model with the Transferable Dialogue State Generator~(TRADE)~\cite{wu2019transferable} as its state tracker model. 
    \item \textbf{HRED + MDBT}: the HRED model with MDBT as its state tracker model. 
%Those are two base dialogue models with different external state trackers, the Globally Conditioned Encoder(GCE)\cite{nouri2018toward} state tracking model has the highest correctness rate in slot discovering. And the Transferable Dialogue State Generator(TRADE)\cite{wu2019transferable}. The dialogue state tracker recognizes the slots from users' utterance and summarize the states as the Seq2seq model's input.

    \item\textbf{HRED + TRADE}: the HRED model with TRADE as its state tracker model.

    \item\textbf{LaRL + TRADE}: the Latent Action Reinforcement Learning~(LaRL)~\cite{zhao2019rethinking} method with TRADE as its state tracker model.
    \item\textbf{HDSA + TRADE}: the Hierarchical Disentangled Self-Attention~(HDSA)~\cite{chen2019semantically} model with TRADE as its state tracker model.
    \item \textbf{HRED-MTSS~(Our model)}: the HRED student model training with a Multiple Teachers Single Student framework.
    \item\textbf{Seq2Seq/HRED/HDSA + Manual states}
Those three comparisons use the same models mentioned above. Instead of the dialogue state extracted by model-based state tracker, we use the human-labelled dialogue states as the model input in the test setting. In a real dialogue situation, there is not human labelling in the user's text. So this setting can be considered an idealized setting to figure out the upper bound performance the models can reach.
\end{itemize}

\begin{table}[!t]
\centering
\begin{tabular}{|cc|ccc|}
     \hline
     \multicolumn{2}{|c|}{\textbf{Distill weights}}& \multicolumn{3}{c|}{\textbf{Multi-domain}} \\
     $\alpha_1$ & $\alpha_2$ & \textbf{BLEU} &\textbf{Inform} & \textbf{Success}  \\
     \hline
     \hline
     0.01 & 0.005 & 17.0 & 71.7 & 63.5 \\
     0.005 & 0.01 & \textbf{18.9} & 73.6 & 61.2 \\
     0.005 & 0.005 & 18.7 & \textbf{77.5} & \textbf{64.9} \\
     0.0025 & 0.005 & 18.1 & 73.1 & 63.9 \\ \hline
     0.01 & 0 & 17.0 & 72.2 & 62.0 \\
     0.005 & 0 & 18.3 & 72.2 & 63.4 \\ \hline
     0 & 0.01 & 18.2 & 77.1 & 64.7 \\
     0 & 0.005 & 18.3 & 74.6 & 63.2 \\ \hline
     0 & 0 & 17.5 & 70.7  & 60.9 \\
     \hline
\end{tabular}
\caption{Results of
adopting different distillation strategies. The last column is the results of a model without distilling.}
\label{tbl:distilpart}
\end{table}

\subsection{Experimental Results and Analysis}  \label{sec:experiments:results}
%\textbf{I believe that if we add more subsubsections to this results subsection, the organization and analysis maybe much clear. -- \textit{FROM SHAOBO}}
\subsubsection{Results on a multi-domain environment} The comparison between our model with the different baseline models is shown in Table~\ref{tbl:multi}. From the table, we can see that compared with the baselines such as the Seq2Seq or the HRED model, our model~(HRED-MTSS) gets the best performance in the multi-domain settings. By adding a teacher-student framework, the informing rate and success rate receive 6.8\% and 4.0\% improvements respectively over the original HRED model. While compared with the state-of-the-art results achieved by HDSA or LaRL with the TRADE state tracker, 
HDSA+TRADE slightly outperforms our model in certain but not all metrics. We have to state that
\begin{itemize}
    \item HDSA uses pre-trained models such as BERT~\cite{devlin2018bert}. However, BERT not only boosts its performance but also brings bloated model and high latency problems in real scenario deployments.
    \item LaRL uses the reinforcement learning method, which aims to maximize the long-term return, i.e., the Inform rate and the Success rate in the dialogue context. LaRL can achieve high scores in one aforementioned metrics but fail in the BLEU score and utterance fluency.
\end{itemize}
Additionally, in the setting of manual states, our model reaches equal or higher results than the Seq2seq and the HRED model. Adding an external state tracker to the Seq2Seq model and the HRED model increases the inform rate but has no help for the dialogue success rate. 
%We think that though the state based on a state tracker has its advantages on representing the dialogue processing, there are still errors exist in the state tracker, implicating the performance of the model. Our model avoids such a problem by using the teacher-student framework.

\subsubsection{Results on single domain environments} As shown in Table~\ref{tbl:domains}, we also test our models' performance in 4 major single domains of MultiWOZ: restaurant, hotel, attraction and train. When compared with a Seq2Seq and HRED model, our model achieves the best success rate in all domains and outperforms in the attraction domain and train domain under the metrics of inform rate . We believe that it is due to the application of an individual teacher in each domain in the training phrase, which results in a better performance in this domain than the universal one. And compared with the HDSA model with the TRADE state tracker, our model is better in 2 of all 4 domains, the restaurant domain and the train domain.

\subsubsection{Individual teachers' performances} We compare the performance between different teachers, a universal multi-domain teacher trained on the whole dataset and the individual teachers trained on respective domains. Table~\ref{tbl:teacher} shows the experimental results of two kinds of teachers in 5 specific domains and 1 generic domain~(The rest 2 domains lack testing data). From the table, we can see that for all domains, the individual teachers get higher BLEU scorers than the universal one. As for the entity matching recall metrics, the individual teachers perform better in 3 of all 5 specific domains. In the restaurant domain, the individual model gets the competitive result over the universal one. The universal model achieves higher entity recall rate than the individual teacher only in the train domain. Results show that the fine-tuned individual teachers significantly outperform the universal model most of the time, while the universal model gets slight advantages only in a few domains. 
We also compare the student's performance with the teachers' and a raw model. Experimental results show that the HRED model applied with MTSS framework, compared with the vanilla HRED model, achieves more satisfying performance in domains whose dataset size is large~(The dataset size of first 5 domains is in a descending order from left to right in Table~\ref{tbl:teacher}).

\subsubsection{Effect of distillation weights} From Table~\ref{tbl:distilpart}, we can see the results of using different guiding weights for text-level~(
$\alpha_1$) and policy-level~($\alpha_2$). Compared with the model without distillation~($\alpha_1=0$, $\alpha_2=0$), text-level distillation~($\alpha_1\neq0$, $\alpha_2=0$) and policy-level distillation~($\alpha_1=0$, $\alpha_2\neq 0$) can bring improvements respectively. Besides, when applied with both distillation methods together with their weights $\alpha_1=0.005$ and $\alpha_2=0.005$, the model gets the highest performance in both the inform rate and the success. Both the two distillation methods help with the student model.

\section{Conclusions} \label{sec:conclusions}
In this paper, we propose a novel approach to building a high-quality multi-domain dialogue system based on a teacher-student framework. We utilize multiple domain-specific teacher models to help a single student model become a multi-domain dialogue expert, which circumvent the knotty multi-domain dialogue state representation problem. To fully take advantage of the knowledge of the teacher models, we creatively make the teacher model impart their knowledge to the student in both text-level and policy-level. 
To discover the potential of the teacher-student framework, we would focus on adopting the framework to the SOTA dialogue models in our future work. 
% In the paper, we propose a multi-domain dialogue generation model trained with a teacher-student framework. The model takes only raw text as input and takes full advantage of the human-labelled states during training. The model behaves better than the one using an external state tracker, with great improvements in the success rate during a conversation.
% The problem exists in our model that it focuses on the text generation during the conversation, and takes no consideration of the knowledge base querying. So our model cannot be regarded as a complete dialogue system. However, we don't th ink it is unable to process. Adding an extra component, such as a memory network can solve the problem.

\section*{Acknowledgments}
This work was supported by the NSFC (No. 61402403), Alibaba Group through Alibaba Innovative Research Program, Alibaba-Zhejiang University Joint Institute of Frontier Technologies, Chinese Knowledge Center for Engineering Sciences and Technology, Engineering Research Center of Digital Library, Ministry of Education, and the Fundamental Research Funds for the Central Universities.
\bibliography{aaai}
\bibliographystyle{aaai}

\end{document}